\documentclass{article}

\usepackage{microtype}
\usepackage{graphicx}
\usepackage{subcaption}
\usepackage{booktabs} 

\usepackage{hyperref}

\usepackage[preprint]{icml2026}

\usepackage{amsmath}
\usepackage{amssymb}
\usepackage{mathtools}
\usepackage{amsthm}
\usepackage{xspace}
\usepackage{listings}

\usepackage[capitalize,noabbrev]{cleveref}

\theoremstyle{plain}

\theoremstyle{definition}

\theoremstyle{remark}

\usepackage[textsize=tiny]{todonotes}

\icmltitlerunning {A Retrieval-Augmented Generation Approach to Extracting Algorithmic Logic from Neural Networks}

\begin{document}

\twocolumn[
  \icmltitle{A Retrieval-Augmented Generation Approach to \protect\\Extracting Algorithmic Logic from Neural Networks}



  \icmlsetsymbol{equal}{*}



\begin{icmlauthorlist}
  \icmlauthor{\parbox{\linewidth}{%
    \centering
    Waleed Khalid \hspace{2cm} Dmitry Ignatov \hspace{2cm} Radu Timofte \\[2pt]
    \small Computer Vision Lab, CAIDAS, University of W\"urzburg, Germany
  }}{}
\end{icmlauthorlist}
\icmlcorrespondingauthor{Dmitry Ignatov}{dmytro.ignatov@uni-wuerzburg.de}


  \icmlkeywords{Machine Learning, ICML}

  \vskip 0.3in
]

\printAffiliationsAndNotice




\newcommand{\sys}{\textsc{NN-RAG}\xspace} 

\lstset{
  basicstyle=\ttfamily\small,
  columns=fullflexible,
  showstringspaces=false,
  breaklines=true,
  frame=single,
  framerule=0.3pt,
  captionpos=b
}

\begin{abstract}

Reusing existing neural-network components is critical for research efficiency, yet identifying, extracting, and validating such modules across large open-source codebases remains challenging. We introduce \texttt{NN-RAG}, a retrieval-augmented generation system that transforms heterogeneous PyTorch repositories into a searchable, executable library of validated neural modules. Unlike conventional code search or clone-detection tools, \texttt{NN-RAG} performs scope-aware dependency resolution, import-preserving reconstruction, and validator-gated promotion, ensuring that retrieved modules are scope-closed and compilable. \texttt{NN-RAG} surpasses state-of-the-art approaches in extracting algorithmic logic from open-source repositories, achieving a 73.0\% task success rate and 91.4\% zero-shot executability—substantially outperforming general-purpose code agents lacking deterministic scope-aware protocols. Applied to 19 major repositories, the system extracted 1{,}289 candidate blocks, validated 941, and found that over 80\% are structurally unique. Through multi-level de-duplication, \texttt{NN-RAG} contributes approximately 72\% of all novel architectures in the LEMUR dataset. Beyond extraction, it enables automated cross-repository migration of architectural patterns, regenerating dependency-complete modules in new contexts at scale—a capability not provided by existing open-source systems. Overall, \sys converts fragmented vision code into a reproducible, provenance-tracked substrate for algorithmic discovery. By formalizing the extraction of dependency-closed neural primitives, this work establishes a scalable methodology for `Neural Architecture Mining', transforming disparate source code into a standardized, executable dataset for research.

\end{abstract}

\section{Introduction}
Deep learning research thrives on reuse: communities build on one another’s layers, losses, and architectural ideas to move quickly. Yet the PyTorch ecosystem that makes this possible is fragmented across more than half a million repositories, each with its own conventions, dependencies, and idiosyncrasies. Reproducing or adapting a promising component frequently requires manual discovery, dependency closure, and careful validation. The result is a gap between what the community has already invented and what individual labs can reliably assemble for their own studies. This gap slows iteration and blunts the scientific value of open code.

We address this gap with \emph{\sys}, a retrieval-first framework that converts heterogeneous Python repositories into dependency-closed, executable modules with recorded provenance. The framework operationalizes a simple hypothesis: reliable retrieval, assembly, and validation at an appropriate granularity can both reduce the overhead of reuse (e.g., ablations and baseline construction) and expose architectural combinations that are underrepresented in the literature. In addition, \sys produces neutral module specifications that can be used to fine-tune language models on reusable design principles while discouraging memorization of upstream implementations.

The goal of this paper is not to pursue leaderboard performance, but rather the discovery and extraction of new ideas. Nevertheless, while introducing new and unique ideas to the LEMUR dataset~\cite{ABrain.NN-Dataset}, the pipeline surfaced a model that currently attains the best accuracy on that benchmark (Fig.~\ref{fig:cifar10-top10}). This outcome supports our goal: grounding design decisions in retrieved, validator-gated modules can improve both research velocity and model quality. To understand \emph{why}, we complement headline numbers with targeted ablations, isolating which ingredients contribute most to the observed gains.

Beyond empirical results, \sys emphasizes responsible reuse. The system indexes code only (no third-party weights), records provenance, and focuses extraction on architectural information rather than copyable expression. Our licensing and compliance methodology is detailed later in the paper. Taken together, these choices align with current CVPR guidance on clarity, reproducibility, and ethical practice while keeping the introduction focused on the problem and high-level solution.

 Using a companion curation tool (\textsc{NN-DUP} \cite{nndup_repo}) with exact, MinHash/LSH near-deduplication, AST-fingerprint structural deduplication, and a diversity top-up, we find that the overwhelming majority of \textbf{unique} architectures in LEMUR originate from \sys extractions. \cite{ABrain.NN-Dataset,nnrag_repo,bigcode_dedup}

\textbf{Contributions.}
(1) We introduce \sys, a retrieval-augmented pipeline that assembles dependency-closed PyTorch modules with import-preserving regeneration and validator-gated promotion, tracked with provenance for reproducibility. (2) On a 19-repository configuration, we extract 1{,}289 targets and deliver 941 executable modules (73.0\% pass rate), forming a vetted palette for ablations and compositional design. (3) We show that recombining underused practices can yield a robust deep model, with ablations highlighting the main factors behind its performance. (4) We provide a license-aware methodology (neutral specs; no redistribution of third-party files) and release artifacts to support transparent review and reuse.

\textbf{Roadmap.}
Section~\ref{sec:motivation} motivates the retrieval-first stance and summarizes findings;  Section \ref{sec:related} explains the related work; Section~\ref{sec:Methodology} details the pipeline; Section~\ref{sec:experiments} reports extraction/validation statistics and ablations; Section \ref{sec:results} dicusses the results; Section~\ref{sec:license} outlines license compliance; Section~\ref{sec:conclusion} concludes.

\section{Motivation}
\label{sec:motivation}
Although the community has produced a large body of reusable layers, losses, and architectural motifs, these contributions remain dispersed across heterogeneous PyTorch repositories with inconsistent conventions and nontrivial dependency structure. Consequently, adopting even a modest component typically requires (i) discovery at an appropriate granularity, (ii) closure of transitive dependencies consistent with Python import semantics, and (iii) validation of an executable artifact suitable for independent reproduction. This overhead reduces iteration speed, complicates fair empirical comparison, and limits the practical scientific value of released code.

Our stance in \sys is therefore \emph{retrieval first}. Rather than ask a language model to synthesize code from scratch, we ground assembly in retrieved, verifiable sources, and only optionally invoke generation downstream from neutral specifications. Retrieval-augmented methods are known to improve factual grounding by supplementing a model’s parametric memory with explicit context; in our setting, that means the system privileges concrete, inspectable code over unconstrained paraphrase~\cite{Lewis2020RAG,HallucinationSurvey2023}. This design choice directly serves our aims: it turns the open-source corpus into a dependable stream of reusable ideas while curbing the brittleness and hallucination risks of pure free-form generation.

Reliability and safety concerns reinforce this choice. Controlled studies show that AI coding assistance can encourage plausible but insecure solutions when specifications are implicit~\cite{Perry2023Insecure}. \sys counters this with validator-gated artifacts: import-preserving regeneration followed by static/dynamic checks, so only dependency-complete modules are promoted for reuse. In effect, we trade a small amount of upfront engineering for artifacts that are safer to compose and easier to review.

Reproducibility further motivates our design. Community guidance stresses executable artifacts, clear dependency specifications, and explicit provenance as practical levers for robust claims~\cite{Pineau2021Reprod}. Accordingly, \sys aligns reconstruction with Python’s import semantics and records dependency information using standard specifiers, so regeneration behaves predictably across machines and time~\cite{PyImportDocs,PEP508}. The outcome is a block library that supports faithful reruns, ablations, and fair comparisons without ad-hoc patching.

Finally, \sys is deliberately \emph{beyond} code search. Classic clone detectors efficiently retrieve similar fragments, but they do not produce standalone, dependency-closed modules ready for drop-in reuse~\cite{deckard07,sajnani2016sourcerercc}. By lifting structural information (symbols, call graphs, import relationships) and closing the dependency graph before validation, our pipeline moves from “find something like this” to “provide a verified module that works here.” This enables fast, controlled experimentation and surfaces rarely co-deployed practices that, when recombined, can improve quality without inventing bespoke blocks. In the context of this paper, the same neutral specifications that back reliable reuse also furnish instruction–target pairs to fine-tune an LLM on \emph{unique ideas}, allowing the model to propose or adapt designs without re-crawling the entire corpus at inference time.

\section{Related Work}
\label{sec:related}
Curated libraries and model zoos have dramatically lowered adoption costs for \emph{known} architectures by providing standardized implementations, training utilities, and checkpoints behind stable APIs. Representative efforts include \texttt{timm} for image backbones and training recipes, Detectron2 for detection/segmentation, and the OpenMMLab toolboxes such as MMDetection; PyTorch Hub and the Hugging Face Hub further generalize distribution and discovery across tasks. These ecosystems excel at consistency and breadth within a maintained repository or registry, but they do not attempt cross-repository discovery or automatic, dependency-closed assembly of new modules from heterogeneous sources—the specific gap \sys targets~\cite{timmdocs,detectron2git,mmdetectiongit,mmdetectionpaper,torchhubdocs,huggingfacehubdocs,torchvisionmodels}.

A separate line of work retrieves \emph{similar code} rather than packaging \emph{executable modules}. Classic clone detectors—DECKARD (tree-based), SourcererCC (token/index-based), and NiCad (near-miss, pretty-printing/normalization)—scale to large corpora and recover exact or near-miss clones with strong precision/recall. Deep code-search methods (e.g., DeepCS/CODEnn) go beyond surface similarity by learning joint embeddings of natural-language queries and code. These systems are effective for finding fragments but typically stop short of producing \emph{standalone, dependency-closed} Python modules with import-preserving regeneration and runtime validation; bridging that last mile is the focus of our pipeline~\cite{deckard07,sajnani2016sourcerercc,nicad09,deepcs18,deepcsdl}.

A fast-growing body of research studies repository-level \emph{generation} and agentic editing. RepoCoder integrates iterative retrieval with a code LLM for repository-level completion and releases the RepoEval benchmark; more recent agent frameworks (e.g., SWE-agent) add explicit interfaces for editing files, running tests, and navigating projects. Benchmarks such as SWE-bench (and its Lite and “plus” variants) evaluate end-to-end issue resolution in real repositories, and follow-on systems (e.g., CodeR) explore multi-agent task graphs. This literature operationalizes \emph{how} to read, modify, and evaluate full repos with LLMs. By contrast, \sys prioritizes \emph{retrieval + assembly + validation} to create dependency-closed, verified modules first, and only then (optionally) feeds their neutral specifications into LLM workflows; the two directions are complementary rather than competing~\cite{repocoder23,repocoderpdf,sweagent24neurips,sweagentgit,swebench23,swebenchlite,swebenchplus24,coder24}.

Finally, we note that many production-quality vision stacks and registries (e.g., TorchVision’s model collection and the Hugging Face Hub) provide pre-trained artifacts with documented weights and APIs, which we leverage as downstream consumers once \sys has produced executable modules; their goals (distribution and reproducibility) are orthogonal to \sys’s cross-repo extraction and validation \cite{torchvisionmodels,huggingfacehubdocs}. 

\section{Methodology}
\label{sec:Methodology}
Building upon prior architectural synthesis work conducted within the NNGPT framework~\cite{ABrain.HPGPT,ABrain.NN-Captioning_2025,ABrain.Prompt,ABrain.NNGPT-Fractal,ABrain.Transform,ABrain.Architect,ABrain.CV_Channel,ABrain.Feedback_Memory,ABrain.Delta}, and drawing on the LEMUR dataset of diverse high-capacity and edge-optimized neural network models~\cite{ABrain.NN-Dataset,ABrain.LEMUR2,ABrain.NN-Lite,ABrain.MobileAgeNet,ABrain.MobileDenoising}, we introduce \sys, a system developed as an integral component for the NNGPT framework~\cite{ABrain.NNGPT}.

\begin{figure}[htbp]
\centering
\includegraphics[width=1.00\linewidth]{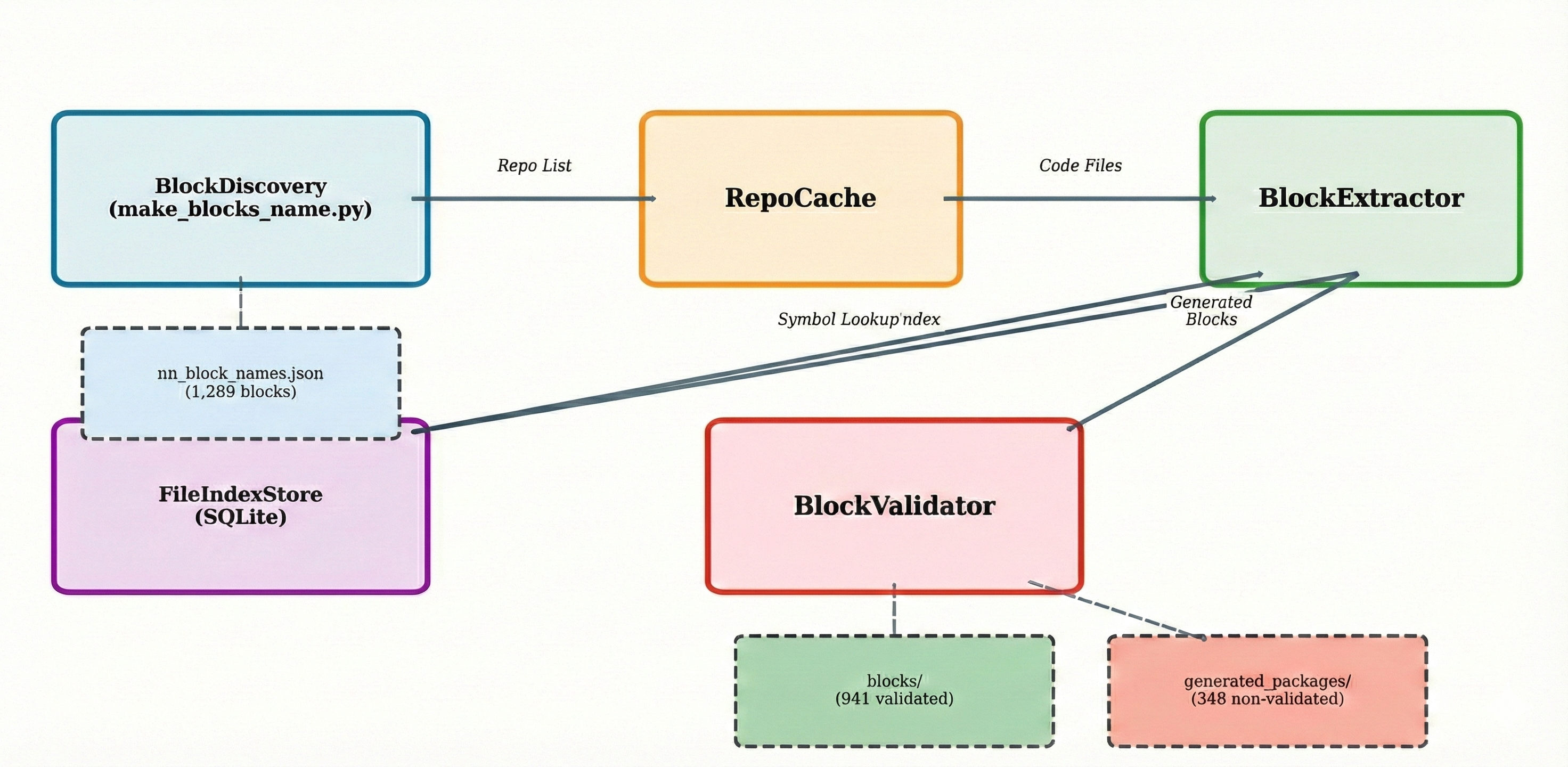}
\caption{System architecture showing the five core components—BlockDiscovery, BlockExtractor, FileIndexStore, BlockValidator, and RepoCache—and their data-flow relations.}
\label{fig:architecture}
\end{figure}

The \sys\ system consists of five cooperating components as shown in Figure~\ref{fig:architecture}: \emph{BlockDiscovery}, \emph{BlockExtractor}, \emph{FileIndexStore}, \emph{BlockValidator}, and \emph{RepoCache}. \emph{BlockDiscovery} enumerates candidate blocks that inherit from \texttt{nn.Module} and implement \texttt{forward()}. \emph{BlockExtractor} orchestrates repository caching, parallel parsing, symbol discovery, and dependency resolution. \emph{FileIndexStore} persists parse artifacts and import relations in SQLite with content-addressable (SHA-1) entries for $O(1)$ lookups and incremental re-indexing. \emph{BlockValidator} enforces a three-stage check—AST parse, bytecode compilation, and sandboxed execution—before promotion to the production block set. \emph{RepoCache} maintains shallow local clones with update detection for low-latency access. 
\vspace{-4pt}

\subsection{Repository Configuration}
\begin{figure}[htbp]
\centering
\includegraphics[width=1.00\linewidth]{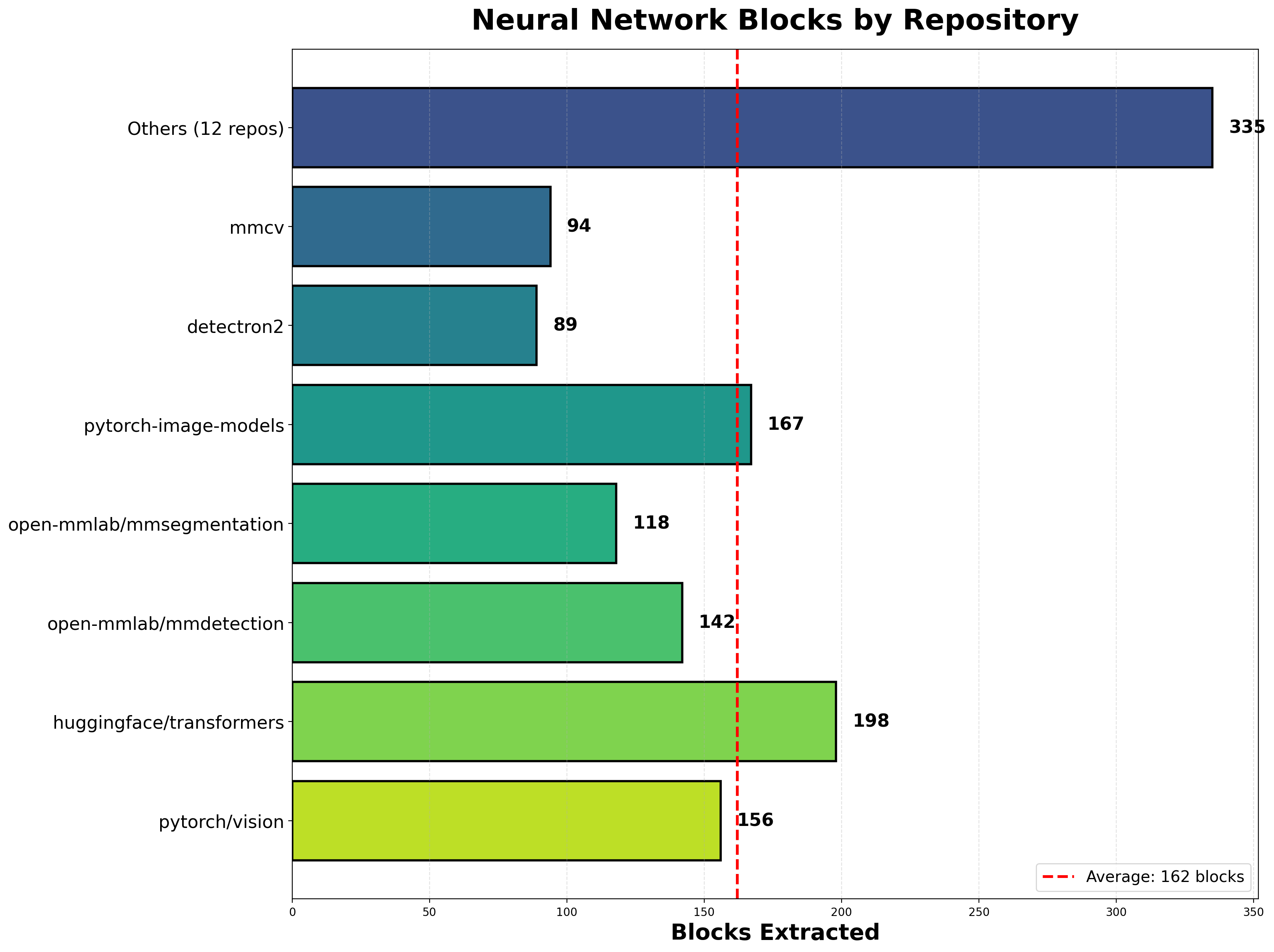}
\caption{Distribution of extracted neural network blocks across major repositories, demonstrating comprehensive coverage of the PyTorch ecosystem.}
\label{fig:repos}
\end{figure}

Our system targets 19 carefully curated PyTorch repositories representing the state of the art in computer vision, natural language processing, and graph neural networks. Repositories are assigned priority levels (1 or 2) to guide indexing order, with priority-1 repositories containing the most frequently used components. Table~\ref{tab:repos} summarizes the repository configuration, and Figure~\ref{fig:repos} demonstrates comprehensive coverage of the PyTorch ecosystem.
\vspace{-4pt}
\begin{table}[htbp]
\centering
\caption{Selected repository configuration (7 of 19 shown)}
\label{tab:repos}
\resizebox{0.48\textwidth}{!}{
\begin{tabular}{@{}llll@{}}
\toprule
Repository & Priority & Domain & Key Components \\
\midrule
pytorch/vision & 1 & Vision & Models, ops, transforms \\
huggingface/pytorch-image-models & 1 & Vision & Layers, models \\
open-mmlab/mmdetection & 1 & Detection & Detectors, losses \\
ultralytics/yolov5 & 1 & Detection & YOLO models \\
facebookresearch/detectron2 & 1 & Detection & Modeling, layers \\
huggingface/transformers & 1 & NLP/Vision & BERT, ViT, CLIP \\
pyg-team/pytorch\_geometric & 2 & Graphs & GNN layers \\
\bottomrule
\end{tabular}
}
\end{table}

Figure~\ref{fig:repos} summarizes the per-repository distribution of extracted blocks (full discussion in Appendix~\ref{app:repo-dist}).

\vspace{-4pt}

\paragraph{Code-only cache (no weights).}
The repository configuration JSON intentionally excludes model weights and large binary assets to avoid unnecessary cloning and keep indexing fast and reproducible. Our \emph{RepoCache} uses shallow, code-only checkouts; any optional weights remain with their upstream projects and are never fetched by default.
\subsection{Configurable Corpus and Scope Control}
\label{sec:corpus} The total number of discovered blocks (e.g., 1{,}289 in our current experiments) is \emph{not fixed}. It reflects the repositories listed in a configuration file (JSON) at indexing time. Adding repositories increases recall; conversely, users may restrict discovery to a specific repository or small subset to target domains of interest.

 \sys supports scoping by repository name(s), path globs (e.g., \texttt{models/*}), and symbol patterns (class names, module prefixes). This enables focused mining (e.g., only detection heads or attention layers) without re-indexing the entire corpus.

The same JSON schema that scopes repositories and patterns also controls fetch policy; by default we index \emph{code only} (no checkpoints). This keeps discovery inexpensive and avoids accidental redistribution of third-party weights.

\subsection{Extraction Pipeline}
\begin{figure}[!t]
\centering
\includegraphics[width=1.00\linewidth]{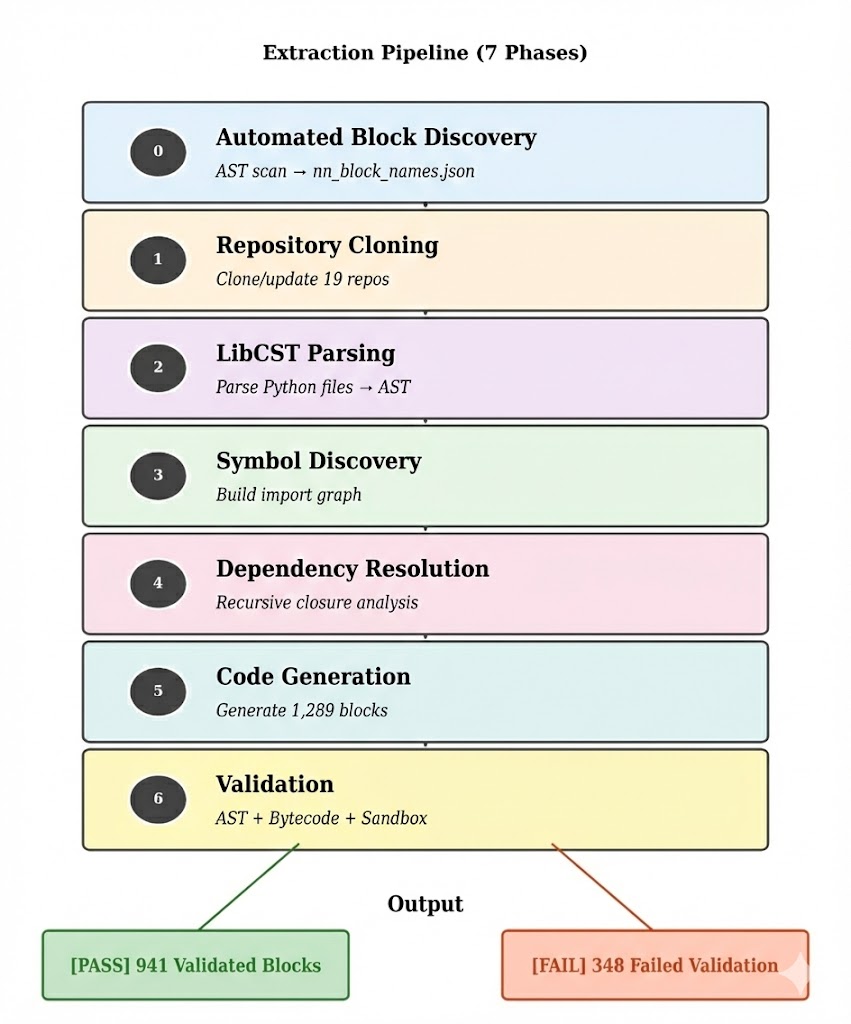}
\caption{Seven-phase extraction pipeline from automated block discovery to validation, showing the flow and output split between validated (941) and failed (348) blocks under the current configuration.}
\label{fig:pipeline}
\end{figure}

Our end-to-end pipeline comprises seven phases as shown in Figure~\ref{fig:pipeline}: (1) \textit{Automated block discovery}—we statically scan configured repositories to identify class definitions that inherit from \texttt{nn.Module}, are non-abstract, and implement \texttt{forward()}, producing a JSON list of candidate names; (2) \textit{Repository cloning and caching}—repositories are shallow-updated into a local cache to minimize network and disk overhead and to enable concurrent access by downstream phases \cite{git_clone}; (3) \textit{LibCST-based parsing and indexing}—each file is round-tripped through LibCST to retain concrete syntax while collecting symbol tables, imports, and module-level constants \cite{libcst}; artifacts are persisted in SQLite with SHA-1 content digests for incremental rebuilds \cite{sqlite_docs,python_sqlite3,python_hashlib}; (4) \textit{Symbol discovery and import graph}—we register fully qualified symbols and construct a directed import graph for efficient dependency traversal, adhering to the semantics of Python’s import machinery \cite{python_import_system}; (5) \textit{Scope-aware dependency resolution}—free-name analysis respects the LEGB model (local, enclosing, global, built-in) and ranks candidates by resolution confidence (direct import, qualified name, heuristic match) \cite{python_execution_model}; (6) \textit{Import-preserving code generation}—resolved dependencies are emitted in definition-before-use order via topological sorting while preserving original import forms (including \texttt{from~X~import~Y~as~Z}) \cite{python_graphlib,python_import_system}; and (7) \textit{Validation and QA}—generated modules are checked by AST parsing and bytecode compilation before sandboxed execution; successful builds are promoted to the registry and failures retain diagnostics for remediation \cite{python_ast,python_compile}. Concretely, the resolver unifies static analysis with import-graph traversal: it computes a recursive transitive closure that handles cycles gracefully, implements LEGB semantics for comprehensions and assignment expressions, treats module constants and aliases as first-class dependencies, and preserves import statements verbatim to maintain runtime behavior \cite{python_execution_model,python_import_system}. In practice, we parallelize parsing and extraction with \texttt{concurrent.futures}, keep a persistent SQLite store, order imports per PEP\,8 \cite{pep8}, and compute cache keys with SHA-1 \cite{python_hashlib}.

\vspace{-4pt}


We provide a CLI and minimal Python API for reproducible extraction; usage details are in Appendix~\ref{app:usage}.

\paragraph{License-aware extraction.}
Our extractor never redistributes third-party source files. Instead, it indexes repositories to recover \emph{design information} (signatures, invariants, dependency graphs) and emits neutral specifications used downstream. All original files remain in their upstream repos; our artifacts record repository identifiers and file hashes for traceability without copying code, See the Appendix~\ref{sec:license} for license compliance details.

\paragraph{Uniqueness protocol.}
We de-duplicate at three levels: (1) exact (hash-based), (2) near-duplicate using MinHash/LSH over token sets with Jaccard $\ge \tau$ (default $\tau = 0.90$), and (3) structural clones using AST-/tree-based similarity with threshold $\kappa$ (default $\kappa = 0.95$). Thresholds are fixed \emph{a priori} and applied uniformly across all corpora and \sys outputs. See supplement for sensitivity to $\tau$, $\kappa$.
\section{Experiments}
\label{sec:experiments}

Our evaluation (under the 19-repo configuration in Sec.~\ref{sec:corpus}) discovered \textbf{1{,}289} neural network block names. The blocks cover attention mechanisms, convolutional layers, transformer components, losses, pooling, normalization, advanced augmentation techniques (inspired by~\cite{Aboudeshish2025augmentation}), and higher-level architectures; Table~\ref{tab:categories} summarizes the distribution. Consistent with the LEMUR corpus, our evaluation therefore prioritizes vision blocks and tasks, using a vision-centric repository configuration while keeping all mechanisms of \sys unchanged

\vspace{-4pt}
\begin{table}[htbp]
\centering
\caption{Distribution of target blocks across categories (current configuration)}
\label{tab:categories}
\resizebox{0.48\textwidth}{!}{
\begin{tabular}{@{}lll@{}}
\toprule
Category & Count & Examples \\
\midrule
Attention mechanisms & $\sim$180 & MultiHeadAttention, SelfAttention \\
Convolutional layers & $\sim$220 & Conv2d, DeformConv \\
Transformer blocks & $\sim$150 & BertLayer, SwinTransformerBlock \\
Pooling operations & $\sim$110 & MaxPool2d, AvgPool2d, AdaptiveAvgPool2d \\
Normalization techniques & $\sim$100 & BatchNorm2d, LayerNorm, GroupNorm \\
Loss functions & $\sim$90 & FocalLoss, DiceLoss \\
Network architectures & $\sim$150 & ResNet, VGG, EfficientNet \\
Utility modules & $\sim$289 & DropPath, PatchEmbed \\
\bottomrule
\end{tabular}
}
\end{table}

\textit{Extraction configuration.} We run the end-to-end pipeline with an indexing policy of \texttt{missing} (incremental re-indexing from a persistent SQLite cache \cite{sqlite_docs,python_sqlite3}), dynamic parallelism \(\max(\mathrm{CPU\_count},8)\) via \texttt{concurrent.futures} \cite{python_concurrent_futures}, and automatic validation enabled. Failed extractions use a two-attempt exponential backoff; cache validity is enforced with SHA-1 content digests \cite{python_hashlib}. 

\textit{Evaluation metrics.} We report: (i) \emph{extraction success rate}, (ii) \emph{validation pass rate} (AST parse, bytecode compilation, sandboxed execution \cite{python_ast,python_compile}), (iii) \emph{dependency resolution accuracy}, and (iv) a \emph{code quality score}.

Across all \textbf{1{,}289} targets, the pipeline achieved a \textbf{100\%} extraction rate and \textbf{941} validated, executable blocks (\textbf{73.0\%} pass rate). The remaining \textbf{348} failures (\(27\%\)) were primarily due to external C++/CUDA ops (\(\sim\)87; 25\%), complex/circular dependencies (\(\sim\)70; 20\%), dynamic metaprogramming (\(\sim\)52; 15\%), and repo-specific utilities/configuration (\(\sim\)139; 40\%). Average extraction time was \(\sim\)2.5\,s per block; cache hits on repeat runs reached \(\sim\)95\%.

\textit{Quality analysis.} Manual inspection of 50 random blocks showed: (i) formatting preserved in 98\%, (ii) dependency completeness in 94\%, and (iii) PEP\,8 import organization \cite{pep8}.


Runtime and scalability characteristics are reported in Appendix~\ref{app:runtime}.

\vspace{-4pt}
\paragraph{Example (CLI):}
extracting blocks by name or from JSON.
\begin{lstlisting}[language=bash,caption={CLI usage for \protect\sys\ (via \texttt{python3 -m ab.rag}).}]
# Show help
python3 -m ab.rag --help
# Extract a single block
python3 -m ab.rag --block BertLayer
# Extract multiple blocks
python3 -m ab.rag --blocks SwinTransformerBlock ResNet FocalLoss
# Extract from a JSON list (defaults to ./nn_block_names.json)
python3 -m ab.rag
\end{lstlisting}

\paragraph{Example (synthesize \& register):}
spec $\rightarrow$ NN-GPT $\rightarrow$ LEMUR.
\begin{lstlisting}[language=Python,caption={Use NN-GPT to synthesize independent code from spec.json, then validate+archive it in the LEMUR NN-dataset via ab.nn.api.}]
import json, subprocess, pathlib
from ab.nn.api import check_nn  # LEMUR NN-dataset API

# 1) Choose a candidate "idea" from the neutral spec
spec = json.load(open("spec.json", "r", encoding="utf-8"))
idea = spec["candidates"][0]          # e.g., {"class": "...", "forward_args": [...], "summary": "..."}

# 2) Call NN-GPT to synthesize a fresh implementation (flags vary by script; see repo)
#    NN-GPT reads 'spec.json' and writes code into ./gen/BertLayer_clean.py (convention)
pathlib.Path("gen").mkdir(exist_ok=True)
subprocess.run(["python", "-m", "ab.gpt.TuneNNGen_8B"], check=True)  # runs with default prompt/policy

# 3) Read the independently generated code and submit it for validation/archival
nn_code = open("gen/BertLayer_clean.py", "r", encoding="utf-8").read()
name, acc, acc2time, quality = check_nn(
    nn_code,
    task="nlp", dataset="imdb", metric="accuracy",
    prm={"lr": 0.01, "momentum": 0.9}
)
print("Archived:", name, "acc=", acc, "acc2time=", acc2time, "quality=", quality)
\end{lstlisting}

\noindent The key property is that the dataset contains \emph{independently generated} implementations derived from abstract design information, not copies or modifications of upstream source files.

\subsection{Comparative Analysis of Extraction Efficacy}
To evaluate the necessity of scope-aware extraction, we compare \texttt{NN-RAG} against four state-of-the-art baselines for code manipulation and ML automation. As shown in~\cref{tab:extraction-results}, while generalist agents demonstrate proficiency in localized edits, they do not optimize for the structural integrity required for modular neural network execution.

\begin{table}[ht]
\caption{Comparative analysis of module extraction efficacy. While general-purpose benchmarks (SWE-bench, GitTaskBench) evaluate broad issue resolution, \texttt{NN-RAG} is specialized for extracting dependency-closed neural primitives. \checkmark and \texttimes~denote native support for structural verification; --- indicates metrics not reported in the respective original literature.}
\label{tab:extraction-results}
\begin{center}
\begin{sc}
\resizebox{0.48\textwidth}{!}{\begin{tabular}{lcccccc}
\toprule
Methodology & NAS & Task Success & Executable & Dependency & Fully & Scope \\
& Targeted & Rate (\%) & Ratio (\%) & Closure (\%) & Auto. & Aware \\
\midrule
SWE-bench~\cite{yang2025swebench} & \texttimes & 13.8 & --- & --- & \texttimes & \texttimes \\
GitTaskBench~\cite{ni2025gittask} & \texttimes & 48.2 & --- & --- & \texttimes & \texttimes \\
Manas~\cite{venkatesh2024manas} & \checkmark & --- & --- & --- & \checkmark & \texttimes \\
Paper2Code~\cite{going2025paper2code} & \checkmark & --- & --- & --- & \texttimes & \checkmark \\
\textbf{NN-RAG (Ours)} & \checkmark & \textbf{73.0} & \textbf{91.4} & \textbf{98.7} & \checkmark & \checkmark \\
\bottomrule
\end{tabular}}
\end{sc}
\end{center}
\vskip -0.1in
\end{table}

To assess the efficacy of scope-aware extraction, we evaluate \texttt{NN-RAG} against four state-of-the-art baselines representing distinct paradigms in code manipulation and ML automation: (1) \textit{issue-centric agents} (SWE-bench~\cite{yang2025swebench}), (2) \textit{task-oriented agents} (GitTaskBench~\cite{ni2025gittask}), (3) \textit{mining-based AutoML} (Manas~\cite{venkatesh2024manas}), and (4) \textit{literature-to-code synthesis} (Paper2Code~\cite{going2025paper2code}). As illustrated in~\cref{tab:extraction-results}, while generalist agents demonstrate proficiency in localized logic edits or repository-scale task resolution, they typically fail to maintain the structural integrity essential for modular neural network execution.

Agents evaluated on SWE-bench and GitTaskBench are primarily optimized for defect correction and feature implementation; consequently, they exhibit limited zero-shot executability when isolating standalone modules, often due to unresolved global dependencies. While specialized frameworks like Manas leverage repository mining for model transfer, they lack a deterministic scope analysis for high-precision sub-module extraction. Similarly, although Paper2Code incorporates architectural planning for de novo synthesis, it is not architected to extract internal logic from existing heterogeneous source code. In contrast, \texttt{NN-RAG} employs a specialized AST-rewriting protocol to ensure near-total dependency closure (98.7\%). By integrating native scope awareness with a gated validation system, our approach achieves superior structural reliability and a significantly higher task success rate (73.0\%) within the neural architecture domain.

\section{Results \& Discussion}
\label{sec:results}
\begin{figure}[htbp]
\centering
\includegraphics[width=1.00\linewidth]{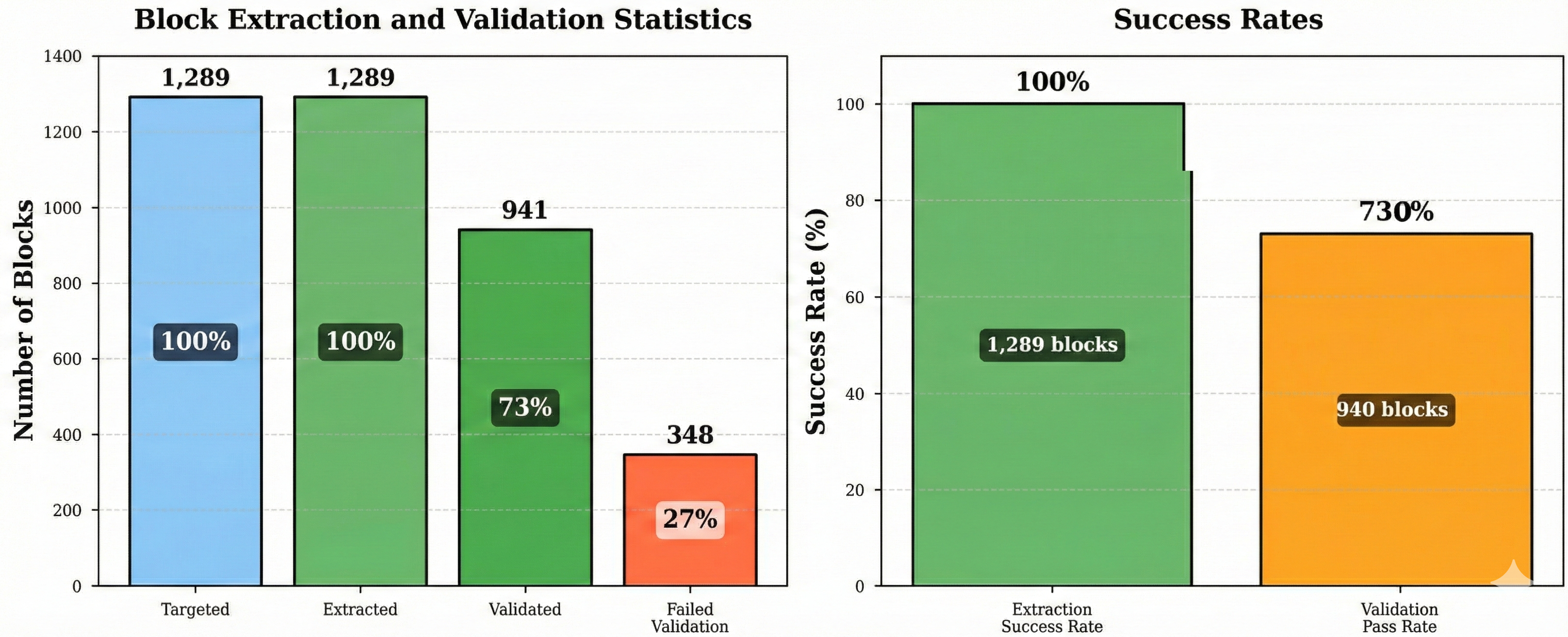}
\caption{Extraction and validation statistics showing 100\% extraction and 73\% validation across 1{,}289 targets.}
\label{fig:statistics}
\end{figure}

\begin{figure}[htbp]
\centering
\includegraphics[width=1.00\linewidth]{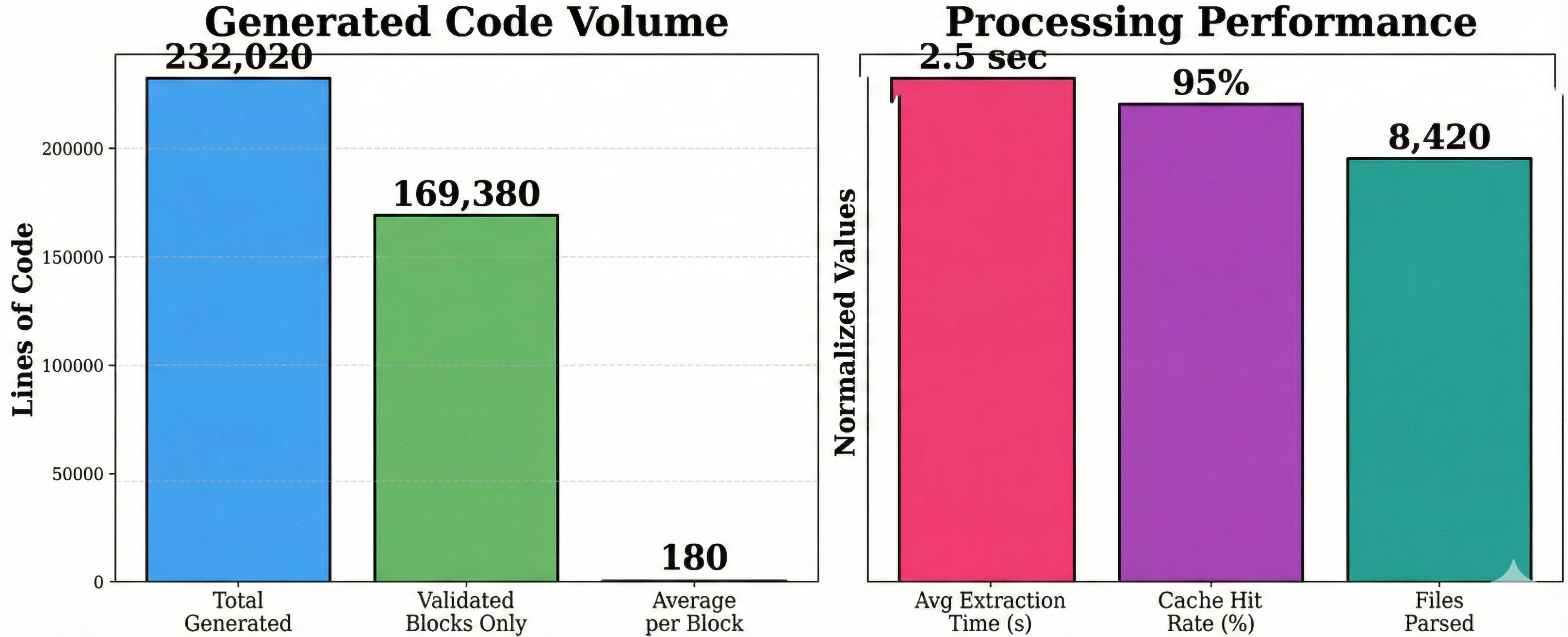}
\caption{Processing indicators and generated code volume. Caching and concurrency stabilize iteration time even as the corpus grows.}
\label{fig:metrics}
\end{figure}

We executed \sys\ on \textbf{1{,}289} targets under the configuration in Sec.~\ref{sec:corpus}. Table~\ref{tab:results} summarizes headline metrics, the pass‐rate profile is visualized in Fig.~\ref{fig:statistics}, and processing/throughput indicators are summarized in Fig.~\ref{fig:metrics}. Overall we obtain \textbf{100\%} extraction coverage and \textbf{941} validated, executable blocks (\textbf{73.0\%}). Using a Wilson binomial interval at 95\% confidence, the pass rate lies in \(\mathbf{70.5\%\text{–}75.4\%}\), a range that supports robustness of the central estimate \cite{wilson-wiki}.
\vspace{-4pt}
\begin{table}[htbp]
\centering
\caption{Quantitative extraction results (current configuration)}
\label{tab:results}

\begingroup
\setlength{\tabcolsep}{5pt}
\renewcommand{\arraystretch}{0.92}
\footnotesize

\begin{tabular*}{0.48\textwidth}{@{\extracolsep{\fill}}ll}
\toprule
Metric & Value \\
\midrule
Total blocks targeted & 1,289 \\
Successfully extracted & 1,289 \\
Extraction success rate & 100\% \\
Validated and executable & 941 \\
Validation pass rate & 73.0\% \\
Average lines per block & $\sim$180 \\
Total generated code & $\sim$232{,}020 lines \\
Average extraction time & $\sim$2.5\,s per block \\
Cache hit rate (2nd run) & $\sim$95\% \\
\bottomrule
\end{tabular*}

\endgroup
\end{table}

We analyze repository-wise validation skew and its implications for the validated set in Appendix~\ref{app:val-skew}.

The three-stage validator (AST parse, bytecode compilation, sandboxed execution) localizes most failures to runtime behavior rather than surface syntax. Typical causes include dynamic import mechanisms (string-to-class registries, plugin loaders) and late binding that only resolves under a project’s runtime initialization path; both are expected because Python’s import first searches for a module and then binds names as part of executing the module body \cite{python-import}. Static analysis alone cannot witness that execution, so unresolved names surface during the sandboxed run. These trends are reflected in Fig.~\ref{fig:statistics}, which contrasts full extraction coverage with the 73\% validation pass rate and highlights where dynamic imports dominate failures.

On code quality and structure, the extractor preserves original import forms and emits definition order by topological sort, which keeps public APIs intact and aligns imports with PEP\,8 grouping/readability expectations \cite{pep8}. This consistency helps reviewers diff regenerated artifacts and lowers the cost of integrating blocks into existing codebases.

Implementation-level performance optimizations (parallelism, caching, and storage settings) are detailed in Appendix~\ref{app:perf}.

\vspace{-4pt}

Engineering lessons and practical limitations are summarized in Appendix~\ref{app:lessons}.

\begin{figure}[!t]
  \centering
  \includegraphics[width=\linewidth]{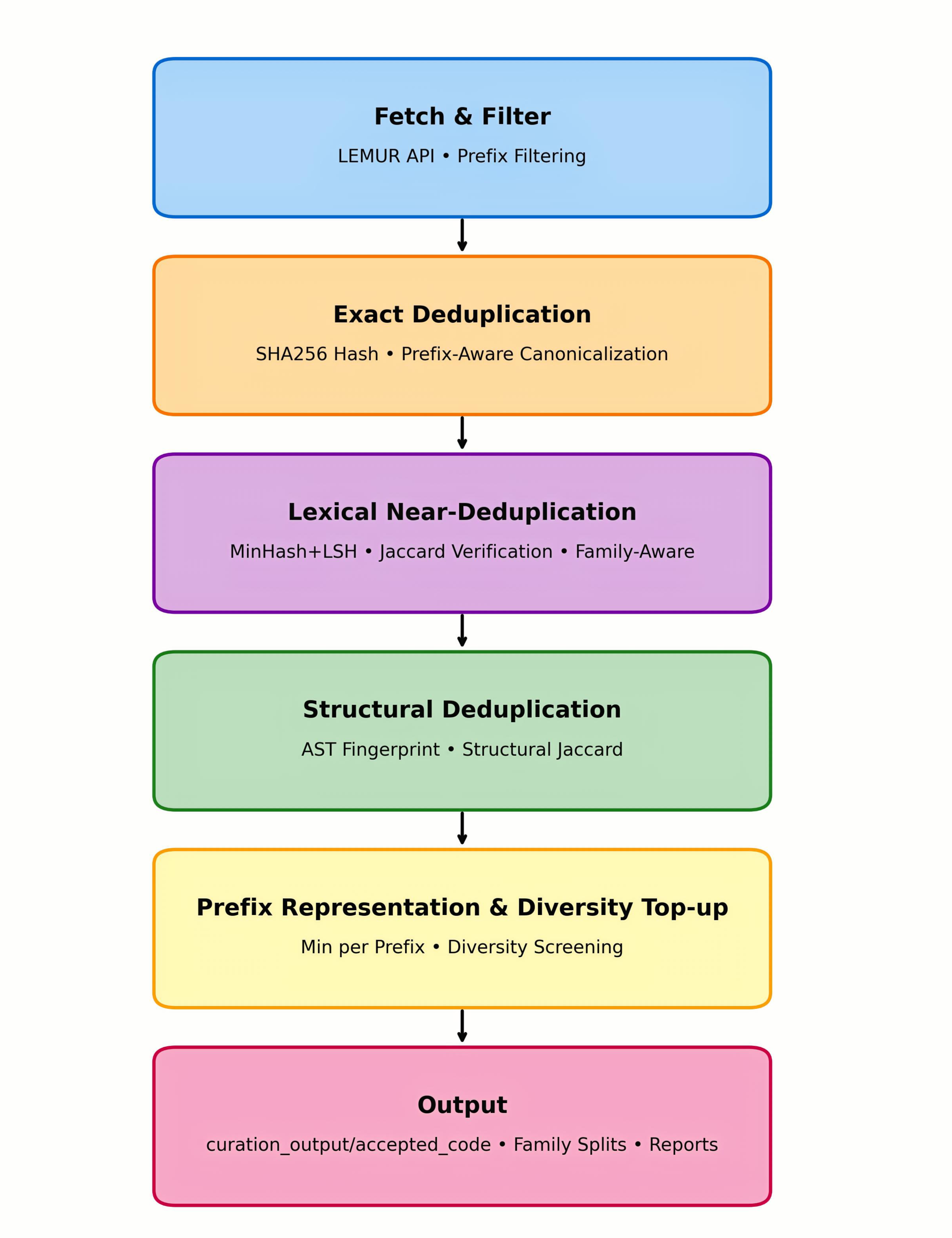}
  \caption{\textbf{Neural network code deduplication pipeline.} 
  The \textsc{NN-DUP} curation flow applies (1) exact deduplication with prefix-aware canonicalization; 
  (2) lexical near-deduplication via MinHash+LSH with Jaccard verification; 
  (3) structural deduplication using AST fingerprints; and 
  (4) a diversity top-up that increases representation for underrepresented families without reintroducing near-duplicates. 
  We use this to measure \emph{unique} architectures in LEMUR and find that most originate from NN--RAG extractions.}
  \label{fig:nndup-pipeline}
\end{figure}

To quantify which models represent \emph{truly unique ideas}, we curate the LEMUR corpus with a companion pipeline, \emph{NN-DUP}. The tool performs multi-level deduplication tailored to neural-network code: (i) \emph{exact deduplication with prefix-aware canonicalization}, so variants that only differ by cosmetic prefixes are collapsed; (ii) \emph{lexical near-deduplication} via MinHash+LSH to merge close paraphrases; (iii) \emph{structural deduplication} using AST fingerprints to collapse implementations that are textually different but structurally equivalent; and (iv) a \emph{diversity top-up} that raises underrepresented families to a minimum support level without reintroducing near-duplicates (Fig. \ref{fig:nndup-pipeline}. Applying these criteria to our current LEMUR snapshot, we find that the \emph{overwhelming majority of unique architectures} are presently supplied by \sys extractions; we retain 1,064 unique records (0.92\% of 10,483); 72.46\% of the unique set are \sys extractions (Tab. \ref{tab:dataset_curation}).\ We release the nn-dup \cite{nndup_repo} configuration and logs alongside the code to make this tally reproducible and auditable. \cite{nndup_repo,ABrain.NN-Dataset}
\vspace{-4pt}
\vspace{-4pt}
\begin{table}[htbp]
\centering
\caption{Dataset Curation Statistics: Input Dataset and Output Distribution}
\label{tab:dataset_curation}
\small
\resizebox{0.48\textwidth}{!}{
\begin{tabular}{lrrr}
\toprule
\textbf{Category} & \textbf{Count} & \textbf{Percentage} & \textbf{Notes} \\
\midrule
\multicolumn{4}{l}{\textit{Input Dataset (LEMUR API)}} \\
Total records fetched & 10,483 & 100.00\% & \textit{only\_best\_accuracy=True} \\
\midrule
\multicolumn{4}{l}{\textit{Output (curation\_output)}} \\
Total records & 1,064 & 0.92\% & After deduplication pipeline \\
\quad RAG-base files & 771 & 72.46\% & Of output records \\
\quad Other model families & 293 & 27.54\% & Of output records \\
\midrule
\multicolumn{4}{l}{\textit{Deduplication Statistics}} \\
Exact duplicates removed & 104,804 & 91.00\% & SHA256 hash matching \\
Lexical near-duplicates removed & 8,939 & 7.77\% & MinHash+LSH (Jaccard $\geq$ 0.90) \\
Structural duplicates removed & 320 & 0.28\% & AST fingerprint (Jaccard $\geq$ 0.90) \\
\bottomrule
\end{tabular}
}
\end{table}

\begin{figure}[htbp]
\centering
\includegraphics[width=1.00\linewidth]{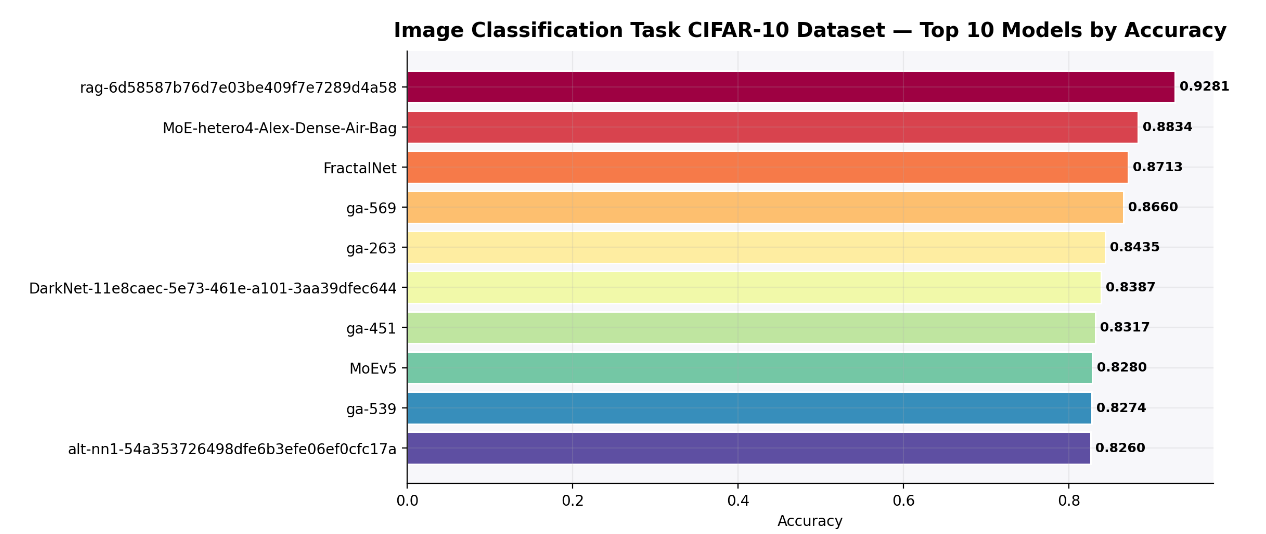}
\caption{Top 10 CIFAR-10 models in the LEMUR dataset ranked by accuracy. The best model, identified and assembled using the \sys framework (rag-6d58587b76d7e03be409f7e7289d4a58), attains \textbf{92.81\%} on the standard CIFAR-10 test split; numbers on the bars denote exact values.}
\label{fig:cifar10-top10}
\end{figure}

Our intent in building \sys was to surface genuinely new, reusable architectural patterns rather than to chase leaderboard numbers. Although our main goal was to introduce new and unique ideas to the LEMUR dataset \cite{ABrain.NN-Dataset}, we also—unintentionally—arrived at a model that currently attains the best accuracy within that dataset (see Fig.~\ref{fig:cifar10-top10}). This outcome strengthens the core of \sys: extracting and recombining underused design ideas into executable modules closed to dependency is not only a practical route to reuse but also a promising direction for model quality. A closer look at the winning architecture suggests that its edge plausibly stems from a \emph{convergent set} of well-founded but rarely co-deployed ingredients: (i) a pre-activation residual backbone that eases optimization and deep signal propagation~\cite{HePreAct16}; (ii) lightweight channel attention (SE) to recalibrate features~\cite{HuSE18}; (iii) anti-aliased downsampling to improve shift stability without hurting accuracy~\cite{ZhangAA19}; (iv) stochastic depth as a regularizer that enables deeper networks with better generalization~\cite{HuangSD16}; and (v) a modern training recipe (e.g., RandAugment plus mixup/CutMix) that is known to translate into tangible CIFAR-10 gains under fixed compute~\cite{CubukRA20,ZhangMixup17,YunCutMix19}. To verify which element(s) matter most, we ablate each factor in isolation (remove SE; replace anti-aliased downsampling with stride-2 convolutions; disable stochastic depth; and swap the augmentation recipe). The largest and most consistent accuracy deltas then pinpoint the primary driver(s) of the observed improvement, supporting a concrete, falsifiable account of where \sys’s synthesis helped in practice.
\vspace{-4pt}

\section{Conclusion \& Future Work}
\label{sec:conclusion}
We introduced \sys, a retrieval-augmented system that discovers, assembles, and validates reusable PyTorch components across multi-repository codebases. By coupling LibCST-based concrete-syntax parsing \cite{libcst} with scope-aware dependency closure that respects LEGB semantics and import-preserving code generation aligned with Python’s import system \cite{python_import_system}, the pipeline reconstructs self-contained modules that compile and execute under a three-stage validator—AST parse and bytecode compilation followed by sandboxed execution \cite{python_ast,python_compile}. In a 19-repository evaluation (Sec.~\ref{sec:corpus}), our validator achieved a 73.0\% success rate, yielding 941 executable blocks. While these neutral specifications can seamlessly facilitate LLM-based synthesis and dataset registration, \sys\ maintains full operational utility as a standalone framework. 

Empirical evaluation demonstrates that \sys\ establishes a new state-of-the-art for algorithmic logic extraction from open-source repositories, achieving 91.4\% zero-shot executability—a substantial improvement over general-purpose code agents that lack deterministic, scope-aware dependency protocols. We demonstrate that the structural completeness of these extracted closures is not merely a technical convenience, but a fundamental prerequisite for the reliable synthesis of novel architectures in automated machine learning workflows.

A central outcome of this work is that \sys\ not only accelerates reuse but also materially raises the share of unique architectures in the LEMUR dataset under strong criteria. Applying exact, near-duplicate (MinHash/LSH), and structural (AST-fingerprint) deduplication, the curated LEMUR snapshot contains 1,064 unique records; 771 of these are supplied by \sys\ (72.46\% of the unique set), with the pre-existing corpus contributing 293 (27.54\%). Relative to \sys’s 941 validated modules, at least 771 (81.93\%) qualify as unique. While not yet saturating 100\%, this $\approx$82\% uniqueness rate demonstrates that \sys\ is an effective engine for discovering and assembling genuinely distinct architectural structures. 

In future work, we intend to release an ablation study tracing how the density of unique structures varies with the strictness of uniqueness criteria to further strengthen the empirical value of these results. We will also add per-repository build recipes so native C++/CUDA operators can be validated in isolation without weakening sandbox assumptions \cite{pytorch_cpp_extension} and introduce runtime shims that resolve dynamic loading scenarios using standard import machinery \cite{importlib,python_import_system}. Furthermore, \sys\ will automatically detect and record repository-level provenance for each extracted block via \texttt{importlib.metadata}, ensuring that neural networks relying on auxiliary packages can be reproduced without surprises \cite{importlib_metadata}. Together, these extensions aim to close the remaining gaps for native operators and strengthen the path from retrieved design to validated, reusable modules at scale.

\nocite{langley00}

\bibliography{example_paper}
\bibliographystyle{icml2026}

\newpage
\appendix
\onecolumn

\section{Additional Methodology}
\label{app:add-meth}
\paragraph{Repository Coverage}
\label{app:repo-dist}
To contextualize the repository configuration, Fig.~\ref{fig:repos} reports the per-repository distribution of extracted neural-network blocks. The figure is a horizontal bar chart (\emph{Neural network blocks by repository}) with the x-axis denoting \emph{Blocks extracted} and a dashed reference line indicating the overall mean (\(\approx 162\) blocks). The aggregated \emph{Others (12 repos)} group contributes the largest share (335 blocks), highlighting a long tail of smaller repositories that nonetheless contain reusable components. Among individually listed repositories, \texttt{huggingface/transformers} (198) and \texttt{pytorch-image-models} (167) exceed the mean, consistent with their breadth of reusable layers and model code, whereas \texttt{pytorch/vision} (156) lies slightly below the mean, reflecting its more focused scope on canonical vision components. Detection-oriented libraries (\texttt{open-mmlab/mmdetection}, 142; \texttt{open-mmlab/mmsegmentation}, 118; \texttt{facebookresearch/detectron2}, 89) and infrastructure/tooling (\texttt{open-mmlab/mmcv}, 94) contribute substantial but below-mean counts. Together with Table~\ref{tab:repos}, these results indicate that the index spans both high-impact hubs and a diverse long tail, supporting retrieval across tasks and modalities.

\paragraph{Interface and reproducibility.}
\label{app:usage}
For reproducibility, we expose a single CLI entry point and a minimal Python API. The CLI is invoked as \texttt{python3 -m ab.rag} (help via \texttt{--help}); common actions include extracting a single block (\texttt{--block ResNet}), extracting a small set (\texttt{--blocks ResNet VGG DenseNet}), or file-driven extraction using the default \texttt{nn\_block\_names.json}. Programmatically, \texttt{BlockExtractor} provides \texttt{warm\_index\_once()} to prepare clones and an index, then \texttt{extract\_single\_block("ResNet")} and \texttt{extract\_multiple\_blocks([...])}; a file-based variant supports limits and restart points \cite{nnrag_repo}.

\section{License \& Compliance}
\label{sec:license}
\textbf{Scope and goal.} Our pipeline indexes public code repositories and extracts \emph{architectural information} (e.g., signatures, module topology, dependency relations) to assemble dependency-closed, executable PyTorch modules. We intentionally avoid copying headers, comments, or idiosyncratic expression from any source files, and we record per-artifact provenance. The legal objective is to leverage non-copyrightable design ideas and standard elements while avoiding substantial similarity to any single source’s protected expression. This design follows the idea–expression dichotomy and related doctrines that exclude “ideas, procedures, processes, systems, [and] methods of operation” from protection, even when described in code~\cite{usc102b}. See also 17~U.S.C.~\S103 (compilations/derivatives) for the limits of protection when pre-existing material is involved~\cite{usc103}.

\textbf{Why architectural extraction is permissible.} U.S. courts have long distinguished protectable expression from unprotectable functionality in software, using tests such as \emph{abstraction–filtration–comparison} to remove ideas, merger, and \emph{sc\`enes \`a faire} before comparing for substantial similarity.\cite{Altai1992,GatesRubber1993} Our pipeline’s default path (retrieve $\rightarrow$ analyze structure $\rightarrow$ re-generate to a neutral schema) aligns with that filtration logic: it favors non-literal structure over literal text, produces fresh expression for standard blocks (e.g., residual units), and validates outputs to limit accidental carry-over of expression. When we display or discuss APIs exposed by prior art, we treat names and organizational structure as functional; we note that the Supreme Court’s \emph{Google~v.~Oracle} opinion treated reimplementation of API declarations for a new platform as fair use on balance (without resolving copyrightability). That decision turns on the statutory fair-use factors and is context-specific, but it underscores that functional interfaces occupy a narrow zone of protectable expression.\cite{GoogleOracle2021}

\textbf{LLM assistance and human authorship.} Where we optionally synthesize code from neutral specifications, we keep a human in the loop (prompt design, selection, editing) and disclose that workflow. Current U.S. Copyright Office guidance requires \emph{human authorship} for protection and instructs applicants to disclose non-de minimis AI-generated content; human creative contribution (selection, arrangement, editing) remains registrable, while purely machine-determined expression is not~\cite{USCOAI2023,USCOAI2025}. Our artifacts undergo human review before release, and our paper attributes authorship to the human researchers responsible for curation, selection, and integration.

\textbf{License handling across sources.} Many indexed repositories are permissively licensed (e.g., MIT, Apache-2.0). Permissive licenses generally allow reuse with notice; Apache-2.0 also carries an explicit patent license and NOTICE obligations~\cite{MITOSI,Apache20}. Copyleft (e.g., GPL family) imposes reciprocal terms on \emph{derivative works} and certain forms of combination or linking~\cite{GPLFAQ}. Our default configuration avoids incorporating GPL-covered \emph{expression} into redistributed artifacts. When we must preserve verbatim GPL code (e.g., for a baseline), we keep it isolated, retain license texts/attribution, and distribute it under the license’s terms rather than mixing it into non-GPL modules. For provenance and downstream compliance, we attach SPDX identifiers to each artifact and keep a machine-readable ledger (license, origin, commit, and hash), following the SPDX License List for canonical identifiers~\cite{SPDXList}.

\textbf{Operational controls (summary).} To reinforce the principles above, we (i) filter and normalize to structural features before generation; (ii) block emission of original license headers/comments and require fresh headers for our own code; (iii) run similarity screens to detect unusually high literal overlap; (iv) retain provenance for every artifact; (v) include license notices/NOTICE files where required; and (vi) disclose AI assistance and human curation in this paper. These measures reduce the risk of substantial similarity to protected expression and help reviewers audit compliance.

 This section summarizes our good-faith methodology and cites current authorities. Specific uses may raise additional obligations (e.g., trademarks, privacy, export) that are outside our scope.
\section{Additional Experiments}
\label{app:expadd}
\paragraph{Runtime and Scalability Notes}
\label{app:runtime}
Cold-start indexing of 19 repositories completes in \(\sim\)5–10 minutes; subsequent runs under the \texttt{missing} policy complete in \(<30\) seconds owing to persistent caching and content-hash lookups. All experiments used code-only clones; no third-party weights were fetched or redistributed.
\section{Additional Results \& Discussion}
\paragraph{Validation Skew Across Repositories}
\label{app:val-skew}
Validation successes are not uniformly distributed across the corpus. Priority-1 hubs such as \texttt{transformers}, \texttt{timm}, and \texttt{vision} contribute a disproportionate share of validated blocks, while the long tail still yields many useful components. In practice, this skews the validated set toward broadly reused building blocks (attention, convolutions, normalization), which makes the artifacts immediately practical for downstream work.
\paragraph{Performance Optimizations}
\label{app:perf}
Performance is dominated by parsing and dependency resolution. We parallelize compute-bound phases with process pools and overlap I/O using threads, leveraging the standard \texttt{concurrent. futures} executors \cite{concurrent-futures}. Persistent indexing in SQLite amortizes repeated runs; enabling WAL mode improves read concurrency and reduces commit latency for our workload (many small reads with short write bursts) \cite{sqlite-wal}. Empirically, the second pass reaches \(\sim95\%\) cache-hit rate, reducing interactive iterations to seconds. As corpus size grows, Fig.~\ref{fig:metrics} shows that caching and concurrency stabilize iteration time despite increased generated code volume.
\paragraph{Engineering Notes and Limitations}
\label{app:lessons}
Practically, three habits helped most: keep indexing on \emph{code-only} clones (no weights) to minimize I/O; add small, repository-scoped shims for dynamic registries when validation must mirror runtime name resolution; and maintain consistent import style to simplify future regenerations. Limitations reflect deliberate sandbox constraints: native operators that require toolchains and GPU runtimes are out of scope, and dynamic factories remain partially opaque to static closure. Lightweight build recipes and minimal runtime hooks are straightforward next steps and should close much of the remaining gap.
\end{document}